\documentclass{article} %

\usepackage{colm2024_conference}

\usepackage{booktabs}
\usepackage{graphicx}
\usepackage{enumitem}
\usepackage{wrapfig}
\usepackage{algorithm}
\usepackage{algpseudocode}

\usepackage{microtype}
\usepackage{amsmath}
\usepackage{colortbl}
\usepackage[utf8]{inputenc}
\definecolor{lightgray}{rgb}{0.9,0.9,0.9}
\usepackage{caption}
\usepackage{subcaption}
\usepackage{xcolor}
\usepackage{setspace}
\usepackage{url}
\usepackage{multirow}
\usepackage{colortbl}
\usepackage{tabularx}
\usepackage{blindtext}
\usepackage{pgfplots}
\pgfplotsset{compat=1.18} 
\usepackage{tikz}
\usetikzlibrary{er,positioning,bayesnet}
\usepackage{makecell}
\usepackage{tipa}
\usepackage{siunitx}
\usepackage{nicefrac}
\usepackage{tocloft}
\usepackage{listings}
\usepackage[raster,skins]{tcolorbox} %
\usepackage{xltabular}
\usepackage{adjustbox}
\usepackage{xurl}
\usepackage{fontawesome}
\usepackage{marvosym}
\usepackage{algpseudocode}

\makeatletter\def\Hy@Warning#1{}\makeatother
\begingroup

\footnotetext{Equal Contribution}
\endgroup

\makeatletter\def\Hy@Warning#1{}\makeatother
\begingroup

\footnotetext{Corresponding authors}
\endgroup

\author{\textbf{Zihan Qiu}$^*$$^{1}$, \textbf{Zeyu Huang}$^*$$^{2}$, \textbf{Bo Zheng}$^*$$^{1}$, \textbf{Kaiyue Wen}$^{3}$, \textbf{Zekun Wang}$^{1}$, \textbf{Rui Men}$^{1}$ \\ \textbf{Ivan Titov}$^{2}$, \textbf{Dayiheng Liu}\textsuperscript{\Letter}$^{1}$, \textbf{Jingren Zhou}$^{1}$, \textbf{Junyang Lin}\textsuperscript{\Letter}$^{1}$
\\$^{1\,}$Qwen Team, Alibaba Group   \,\,$^{2\,}$University of Edinburgh  \,\,$^{3\,}$Stanford University}

\title{Demons in the Detail: On Implementing Load Balancing Loss for Training Specialized Mixture-of-Expert Models}

\begin{document}

\maketitle

\begin{abstract}
This paper revisits the implementation of \textbf{L}oad-\textbf{b}alancing \textbf{L}oss (LBL) when training Mixture-of-Experts (MoEs) models. Specifically, LBL for MoEs is defined as $N_E \sum_{i=1}^{N_E} f_ip_i$, where $N_E$ is the total number of experts, $f_i$ represents the frequency of expert $i$ being selected, and $p_i$ denotes the average gating score of the expert $i$. 
Existing MoE training frameworks usually employ the parallel training strategy so that $f_i$ and the LBL are calculated within a \textbf{micro-batch} and then averaged across parallel groups.
In essence, a micro-batch for training billion-scale LLMs normally contains very few sequences. 
So, the micro-batch LBL is almost at the sequence level, and the router is pushed to distribute the token evenly within each sequence.
Under this strict constraint, even tokens from a domain-specific sequence (\textit{e.g.}, code) are uniformly routed to all experts, 
thereby inhibiting expert specialization.
In this work, we propose calculating LBL using a \textbf{global-batch} to loose this constraint. 
Because a global-batch contains much more diverse sequences than a micro-batch, which will encourage load balance at the corpus level. 
Specifically, we introduce an extra communication step to synchronize $f_i$ across micro-batches and then use it to calculate the LBL.
Through experiments on training MoEs-based LLMs (up to \textbf{42.8B} total parameters and \textbf{400B} tokens), we surprisingly find that the global-batch LBL strategy yields excellent performance gains in both pre-training perplexity and downstream tasks.
Our analysis reveals that the global-batch LBL also greatly improves the domain specialization of MoE experts.
\end{abstract}

\section{Introduction}

In recent years, the Mixture-of-Experts (MoE) framework~\citep{DBLP:conf/icnn/SzymanskiL93, DBLP:conf/iclr/ShazeerMMDLHD17} has become a popular technique to scale the model parameters up~\citep{jiang2024mixtral, dai2024deepseekmoe, liu2024deepseek, yang2024qwen2}.
For instance, Mixtral-8x22B~\citep{jiang2024mixtral} (141B), Deepseek-v3~\citep{liu2024deepseek} (671B) and MiniMax-01~\citep{li2025minimax} (456B) reach a scale of hundreds of billion parameters while maintaining affordable training and inference efficiency.
Typically, standard MoE comprises a \textit{router} network and a group of parallel \textit{expert} modules. Given a set of inputs, the \textit{router} distributes each input to its corresponding experts conditionally and sparsely. Then, the outputs from individual experts are aggregated based on the importance weight that the router assigned to the expert.

One critical factor for training MoE-based models is encouraging the router to assign input to experts in a balanced manner~\citep{DBLP:conf/iclr/ShazeerMMDLHD17, DBLP:journals/jmlr/FedusZS22, zoph2022st,qiu2024layerwiserecurrentroutermixtureofexperts}.
The reasons are twofold: (1) \textit{effectiveness}: if the router continually prioritizes some experts during training, these experts will get more updates than others and will soon dominate that MoE layer, finally resulting in parameter redundancy issue~\citep{DBLP:conf/iclr/ShazeerMMDLHD17, wang2024auxiliary};
(2) \textit{efficiency}: 
training and deploying large-scale MoE-based models often requires the \textit{Expert Parallel}, where different experts will be in different parallel groups to process their inputs. 
Then, their outputs will be gathered and aggregated. 
In this case, the imbalanced expert utilization would heavily slow the forward process.
Therefore, previous works training MoE-based LLMs generally employ an auxiliary loss, called Load-balancing Loss (LBL), to encourage the balanced routing decision~\citep{DBLP:conf/iclr/ShazeerMMDLHD17}.

\begin{figure*}[t!]
    \vskip  -0.2in
    \begin{center}\centerline{\includegraphics[width=1.0\textwidth]{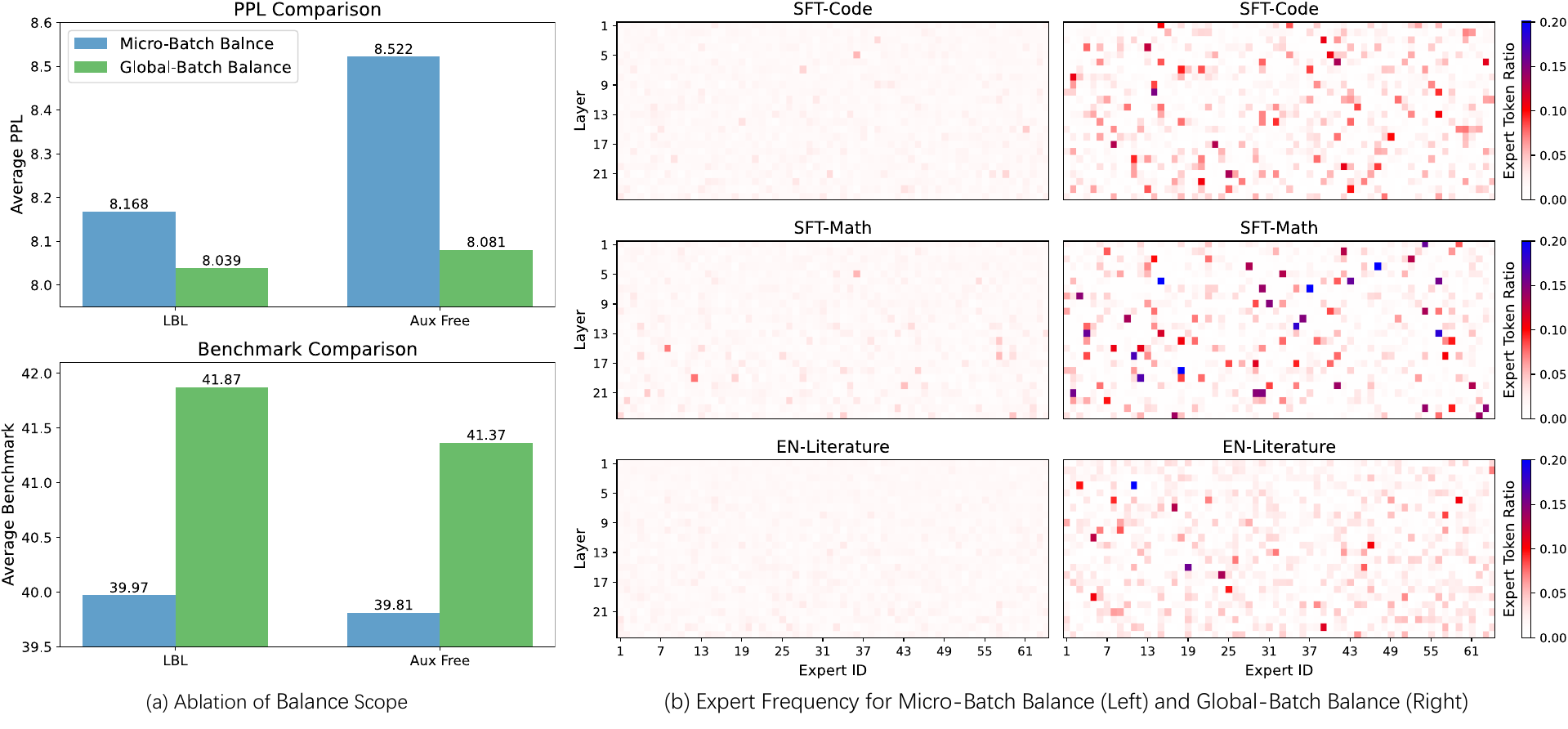}}
\vskip  -0.1in
  \caption{\footnotesize The impact of the Balance BSZ on \textbf{(a)} model performance and \textbf{(b)} expert specialization. 
  \textbf{(a)} When employing micro-batch level load balance, methods based on LBL and based on auxiliary-loss-free~\citep{wang2024auxiliary} approaches perform worse than employing the global-batch balance. 
  \textbf{(b)} When employing micro-batch balance, there is no significant difference in the selection frequency of different domain-specific data, and the selection frequency of different experts within the same domain is approximately the same. 
  With global-batch balance, there is a noticeable difference in the selection frequency of experts on different domain data, and within the same domain, there are experts with high selection frequency (marked in \textcolor{blue}{blue}).
  }
  \label{fig:Specialize}
  \end{center}
  \vskip  -0.3in
\end{figure*}

Nevertheless, in most open-source MoE training frameworks like Deepspeed-MoE~\citep{liu2024deepseek}, Tutel~\citep{hwang2023tutel}, Megablocks~\citep{gale2023megablocks} and Megatron-Core~\citep{shoeybi2019megatron}, the LBL is calculated at the \textit{micro-batch level}, which, as we will soon empirically demonstrate, negatively affects the performance and expert specialization of MoE-based LLMs.
Specifically, during large-scale MoE training, each micro-batch usually contains only up to thousands of tokens and, thus, only a handful of sequences.
Therefore, the micro-batch LBL is almost calculated at the \textit{sequence level}. 
Suppose a micro-batch contains some domain-specific sequences (\textit{i.e.} code and math); the micro-batch LBL still pushes routers to distribute these domain-specific tokens to all experts evenly, introducing an overly strict constraint and may hurt the model performance.

Consequently, we propose calculating the LBL at the \textit{global-batch} level by synchronizing the expert selection frequency across parallel groups and then computing the LBL.
According to the Fig.~\ref{fig:Specialize} (a), the global-batch LBL significantly enhances model performance (\textbf{approximately 0.1 in pre-training PPL and 2 in benchmark scores}).
Additionally, Fig.~\ref{fig:Specialize} (b) showcases that the domain specialization only clearly emerges when trained with the global-batch LBL. 
We demonstrate that the model performance effectively increases with the global batch size (Section~\ref{sec:main_results}).
We conducted further ablation studies to verify that introducing more diverse training tokens instead of more training token numbers is the main contributor to performance gains (Section~\ref{sec:analysis}).
Because the expert selection frequency is just an expert-number-dimensional vector, our method introduces less than 3\% latency under appropriate configurations and achieves more performant and interpretable models.

In summary, we investigate the challenges associated with the LBL in training MoEs models.
By introducing global-batch LBL, we achieve improved performance and foster expert specialization within the MoE model. 
We believe this advancement addresses an essential limitation in existing MoE training, offering a novel perspective for MoEs model optimization. 
Though mainly experimenting with language-based tasks, we hope our work could pave the way for training stronger and more specialised MoE models in various domains, \textit{e.g.}, Computer Vision and Multi-Modality.

\section{Preliminary}

\subsection{Mixture-of-Experts}

MoEs consist of several parallel modules (the `experts') and a router that assigns weights to each expert for a given input.~\citep{DBLP:conf/icnn/SzymanskiL93, DBLP:conf/iclr/ShazeerMMDLHD17} 
Combined with the transformer layer~\citep{vaswani2017attention}, the most common approach is to introduce a set of parallel feed-forward networks (FFN).
Suppose there are $ N_E $ experts, denoted as $E_i, i \in [1, N_E]$. The router $g$ followed by a $\operatorname{softmax}$ function maps the input $\mathbf{x}$ to a score distribution over the experts, $ \operatorname{softmax}(g(\mathbf{x})) \in \mathbb{R}^{N_E} $. 
Typically, for each input, only topK experts with the highest scores are activated and used.
Given $\mathbf{x} \in \mathbb{R}^h$, the output $\mathbf{y} \in \mathbb{R}^h$ is the weighted sum of the outputs from all experts:
\begin{equation}
\label{eqn:MoE}
\mathbf{y} = \sum_{i \in N_E, \, g_i \in \operatorname{topK}(g)} g_i(\mathbf{x}) E_i(\mathbf{x})
\end{equation}

\subsection{Load-balancing Loss}
The Load-balancing Loss (LBL) in training MoE models is a regularization technique that encourages balanced expert utilization and prevents expert collapse~\citep{DBLP:journals/jmlr/FedusZS22}.
Without the LBL, the model tends to concentrate its updates on a limited subset of experts, leading to a severe imbalance in expert utilization. 
To address this issue, LBL penalizes the router if it routes excessive tokens to a few particular experts.
To compute LBL for a batch of tokens, we consider the fraction of tokens 
$f_i$ routed to each expert $E_i$ and the total routing probability $P_i$
allocated to the expert $E_i$. 
The LBL is calculated as the sum of the product of $f_i$ and $P_i$ across all experts $N_E$, normalized by the number of experts:
\vskip  -0.1in
\begin{equation}
\text{LBL} = N_E \sum_{i=1}^{N_E} f_i \cdot P_i.
\label{eq:LBL}
\end{equation}
\vskip  -0.1in
By minimizing the load-balancing loss, the model is encouraged to distribute the considered tokens more evenly among the experts, ensuring that each expert receives a fair share of updates during training. 
This helps maintain a balanced utilization of experts and prevents the model from collapsing into only activating just a few experts.

However, when employing data parallelism and model parallelism strategies for LLMs pre-training, each parallel group (\textit{i.e.}, one GPU) only has data from very limited domains. 
Existing MoE frameworks~\citep{shoeybi2019megatron, gale2023megablocks} only utilize the information of $P_i$ and $F_i$ within every single parallel group when calculating the LBL and then perform an all-gather operation to average:
\begin{equation}
\text{LBL}_{\text{micro}} = \frac{1}{N_P} \sum_{j=1}^{N_p}(N_E \sum_{i=1}^{N_E} f^j_i \cdot P^j_i).
\label{eq:LBL-micro}
\end{equation}
$N_P$ is the number of parallel groups and $f^j_i, P_i^j$ are the fraction and probability in parallel state $j$.
This loss requires the model to \textit{achieve load balance within each parallel group}, thus we call it $\text{LBL}_{\text{micro}}$.
However, suppose one parallel group (or one micro-batch) contains the domain-specific sequences. 
In that case, the router is still pushed to distribute tokens uniformly to all experts, thereby hurting MoE performance and preventing specialization.
This situation is even more common regarding LLMs pretraining. 
Because to better control the data diversity, one micro-batch is usually formed with packed and truncated sequences from one specific domain, and a global-batch consists of micro-batches sampled from different domains according to specific data recipes~\citep{ding2024fewer, yang2024qwen2}.
% Since each parallel group only contains a handful of sequences from one specific domain like code or math, $\text{LBL}_{\text{micro}}$ will push routers to distribute input tokens evenly within every domain.
\textit{So the micro-batch-based balancing approach will hinder the MoE model from allocating data from specific domains to certain experts}, 
which also partially explains why most MoE models only observe token-level expert routing patterns rather than expert-level selections.~\citep{jiang2024mixtral, xue2024openmoe}.

\section{Method}

This section introduces how to turn the micro-batch LBL into global-batch LBL by allowing different parallel groups to synchronize their expert select frequencies. 
We then discuss the scenario in which the number of compute nodes is limited and the sum of micro-batches is smaller than the global batch size. 
In such cases, we propose using a buffer to store the synchronized expert select counts at each gradient accumulation (GA) step to approximate the global batch LBL.

\paragraph{Synchronizing expert selection frequency across parallel groups.}
\label{sec:sync}
Thanks to the format of the LBL calculation in each parallel group, as shown in equation \ref{eq:LBL-micro}, we can synchronize $f_i$ across all parallel groups to get the $\bar{f_i}$ for the global batch. This allows the final global average LBL to be equivalent to the LBL computed by aggregating statistics from all tokens in the global-batch.
\vskip  -0.1in
\vskip  -0.1in
\begin{align}
    \text{LBL}_{\text{global}} &= N_E \sum_{i=1}^{N_E} \bar{f_i} \cdot \bar{P_i} \\ \label{eq:LBL-global}
    &= N_E \sum_{i=1}^{N_E} \bar{f_i} \cdot (\frac{1}{N_P} \sum_{j=1}^{N_p}P_j) \\
    &= \frac{1}{N_P} \sum_{j=1}^{N_p}(N_E \sum_{i=1}^{N_E} \bar{f_i} \cdot P^j_i)
\end{align}
Communicating only $f_i \in \mathbb{R}^{N_E} $ avoids the communication overhead of directly transmitting the token-expert selection matrix of each parallel group and the expert selection scores (with a shape of tokens numbers $\times$ experts numbers).
We develop this synchronization method independently and existing work GRIN~\citep{liu2024gringradientinformedmoe} also proposes this synchronization. 
We will discuss this in the related work part.

\begin{algorithm}
\caption{Approximate Global-Batch LBL}
\begin{algorithmic}[1]
\State Initialize an empty buffer for each expert, $c_i=0$
\While{training continues}
    \For{each gradient accumulation step}
        \State Add $c_i$ with new synchronized selection counts for expert $i$
        \State Calculate the current $f_i$ with $c_i, i \in N_E$ in the buffer
    \EndFor
    \State Optimizer step, clear gradient
    \State Reset the buffer with $c_i=0$
\EndWhile
\end{algorithmic}
\label{algo:apprix_global_batch}
\end{algorithm}
\vskip  -0.1in

\paragraph{Using a buffer to approximate the Global-Batch LBL.} 
\label{sec:buffer}
However, when training LLMs, the global-batch size is often up to $10^3$.
When each micro-batch size is less than $10^1$, due to the limited number of compute nodes, the sum of all micro-batch sizes is smaller than the global-batch size, thus gradient accumulation (GA) is often used.
In this situation, we introduce a buffer to store synchronized $c_i$, the expert $i$'s selection count across micro-batches in one gradient accumulation step. 
Then, the information in the buffer is used to calculate the current $f_i$ at each GA step.
After completing the GA, the buffer is reset. The complete algorithm is shown in the Alg.~\ref{algo:apprix_global_batch}. 
Through this accumulation process, $f_i$ approaches $\bar{f_i}$ with gradient accumulation steps, approximating $\text{LBL}_{\text{global}}$ with limited compute nodes.
\vskip  -0.1in

\section{Experiments}

\subsection{Experimental Setups}

\paragraph{Model Architecture and Training Settings} 
We conduct experiments on three sizes of MoE models: \textbf{(1)} 3.4B total parameters with 0.6B activated (\textbf{3.4A0.6B}); \textbf{(2)} 15B total parameters with 2.54B activated (\textbf{15A2.54B}), and \textbf{(3)}43B total parameters with 6.6B activated (\textbf{43A6.6B}).
Each model utilizes the fine-grained expert~\citep{dai2024deepseekmoe} and shared experts~\citep{rajbhandari2022deepspeed, dai2024deepseekmoe} methods. 
Specifically, the 3.4A0.6B model employs 64 total experts with top4 activated and 4 shared experts, while the 15A2.54B and 43A6.6B models use a setting of 160 total experts with top4 activated and 4 shared experts. 
All models default to using softmax gating, micro-batch LBL, and z-loss.
The auxiliary loss weights follow previous works~\citep{zoph2022st}.
\textit{To avoid the impact of token drop for different methods, we use the dropless routing strategy like dMoE~\citep{gale2023megablocks}.}
In the 3.4A0.6B setting, we also implement the auxiliary loss free (with sigmoid gating) method~\citep{wang2024auxiliary}. 
We train the models on 120B and 400B high-quality tokens, encompassing multilingual, math, and general knowledge content. 
A sequence length of 4096 is used, with global-batch sizes of 512 and 1024 for the 120B and 400B training settings, respectively, comprising 60k and 100k optimization steps. 
% All models adopt a scheduler of warming up 2k steps and cosine decaying to 3e-5. 
% The 3.4A0.6B model uses max lr of 3.2e-3 and 3.83e-3 for the 120B and 400B training settings, respectively. 
% The 15A2.54B model uses a max lr of 1.67e-3 for the 400B training setting, and the 43A6.6B model uses 7.75e-4 for the 120B training setting. 
Other hyperparameters follow the default values of the AdamW optimizer.
We use the term \textbf{Balance BSZ} to indicate the number of tokens considered when calculating the expert selection frequency.

\paragraph{Evaluation} We mainly test the zero-shot downstream task capabilities of the models on four popular benchmarks, including English, Hellaswag~\citep{zellers2019hellaswag}, general knowledge MMLU~\citep{hendrycks2020measuring}, math GSM8k~\citep{cobbe2021training}, and Chinese proficiency C-eval~\citep{huang2024c}. 
Given that benchmarks that are evaluated with accuracy have certain random factors, for more detailed analysis, we mainly refer to the PPL on held-out test sets, which include SFT-EN, EN-Literature, SFT-Code, SFT-Math, SFT-ZH, ZH-Law, ZH-Literature, and SFT-Other from different domains.

\begin{table}[t!]
\centering
\vskip  -0.25in
\caption{\footnotesize Performance of different balance methods and Balance BSZ. 
`LBL' refers to using LBL, and Aux Free refers to the auxiliary loss free method~\citep{wang2024auxiliary}. 
`LBL+sync' means synchronizing expert selection frequency across 
parallel groups in~\ref{sec:sync}. 
`LBL+sync' means further using a buffer to expand the Balance BSZ in~\ref{sec:buffer}.}
\vskip  -0.1in
\label{tab:main-results}
\resizebox{0.8\textwidth}{!}{
\begin{tabular}{cc|ccccc}
\toprule
\textbf{Balance Method} & \textbf{Balance BSZ} & \textbf{Hellaswag} & \textbf{MMLU} & \textbf{GSM8k} & \textbf{C-eval} & \textbf{Avg PPL} \\
\midrule
\multicolumn{7}{c}{MoE-3.4A0.6B (Train 120B Tokens, Global Batch Size 512)} \\
\midrule
LBL & 4 & 62.81 & 41.63 & 13.57 & 41.87 & 8.167 \\
LBL+sync & 32 & 63.58 & 42.08 & 15.01 & 41.58 & 8.062 \\
LBL+sync & 512 & \textbf{63.75} & \textbf{43.48} & \textbf{15.31} & 44.95 & \textbf{8.038} \\
Aux Free & 4 & 61.99 & 41.30 & 12.43 & 43.53 & 8.521 \\
Aux Free & 512 & 63.51 & 42.74 & 14.18 & \textbf{45.03} & 8.080 \\
\midrule
\multicolumn{7}{c}{MoE-3.4A0.6B (Train 400B Tokens, Global Batch Size 1024)} \\
\midrule
LBL & 4 & 67.21 & 48.97 & 21.30 & 49.02 & 7.347 \\
LBL+sync & 128 & 68.08 & 49.02 & \textbf{28.81} & 49.12 & 7.214 \\
LBL+sync & 512 & \textbf{68.32} & 49.84 & 25.40 & \textbf{51.59} & \textbf{7.198} \\
LBL+buffer & 128 & 68.18 & \textbf{49.59} & 24.94 & 50.37 & \textbf{7.199} \\
\midrule
\multicolumn{7}{c}{MoE-15A2.54B (Train 400B Tokens, Global Batch Size 1024)} \\
\midrule
LBL & 16 & 75.69 & 59.99 & 48.07 & 64.38 & 5.778 \\
LBL+sync & 512 & \textbf{76.96} & \textbf{60.78} & \textbf{54.28} & \textbf{64.31} & \textbf{5.603} \\
\midrule
\multicolumn{7}{c}{MoE-43A6.6B  (Train 120B Tokens, Global Batch Size 512)} \\
\midrule
LBL & 8 & 75.2 & 54.98 & 42.08 & 57.06 & 5.862 \\
LBL+buffer & 128 & \textbf{75.94} & \textbf{57.30} & \textbf{46.32} & \textbf{57.98} & \textbf{5.779} \\
\bottomrule
\end{tabular}
}
\vskip  -0.15in
\end{table}

\subsection{Main Results}
\label{sec:main_results}
\paragraph{Global load balance boosts model performance.} 
In this section, we compare using micro-batch and global-batch balances on three different sizes of MoE models under various training scales. 
The 3.4A0.6 B model can be trained using only data parallelism, with a maximum micro-batch size of 4. 
If $f_i$ is synchronized among the 8 GPUs on the same node, the Balance BSZ can reach 32.
When training with 16 nodes and synchronizing across data parallel groups, the Balance BSZ can reach 512. 
From the first part of Tab.~\ref{tab:main-results}, it can be seen that \textit{as the Balance BSZ increases, all metrics consistently improve}. 
For the aux-free method, we also compare the results under micro-batch and global-batch conditions and find the latter is much more better.
For the 3.4A0.6B model trained on 400B tokens, we compare the results when the Balance BSZ could only reach 128 due to the limited number of compute nodes with the results when a buffer is added to approximate the global-batch. 
The latter's performance is closer to the results with a Balance BSZ of 512 and significantly better than 128, proving that introducing a buffer can approximate the global-batch when nodes are limited. 
As training the 15A2.54B and 43A6.6B models requires using model parallelism strategies, we employ expert parallelism for both models, allowing a micro-batch size of 2 and 1 per GPU, respectively. 
We compared the results of synchronizing $f_i$ within the same machine and across all data parallel groups, as shown in the last two parts of Tab.~\ref{tab:main-results}. 
It is evident that increasing the Balance BSZ also significantly improves larger models.

\paragraph{Global load balance encourages expert specialization.} 
We further analyse the selection frequency of each layer's experts across different domains using the held-out PPL test data. 
Specifically, for a given domain, we record the topK experts chosen by each token and calculate the frequency of each expert selection. 
In Fig.~\ref{fig:Specialize}, we compare the expert selection distributions under SFT-Code, SFT-Math, and EN-Literature for models trained with micro-batch balance and global-batch balance. 
It can be observed that (1) with micro-batch balance, most of the selection frequency is the same under EN-Literature, and only a few experts have slightly higher frequencies under SFT-Code and SFT-Math, yet none exceed 0.15. 
This aligns with existing analysis about MoE specialization: models using default load balance hardly exhibit domain-level specialization and only show some token-level specialization~\citep{jiang2024mixtral, xue2024openmoe}. 
(2) In contrast, with global-batch balance, more pronounced high-frequency experts emerge, with many experts in SFT-Math having frequencies exceeding 0.2. 
This confirms our previous discussion that global-batch balance is more conducive to domain specialization.

\begin{wrapfigure}{r}{0.5\textwidth}
\begin{center}
\vskip  -0.2in
\includegraphics[width=0.5\textwidth]{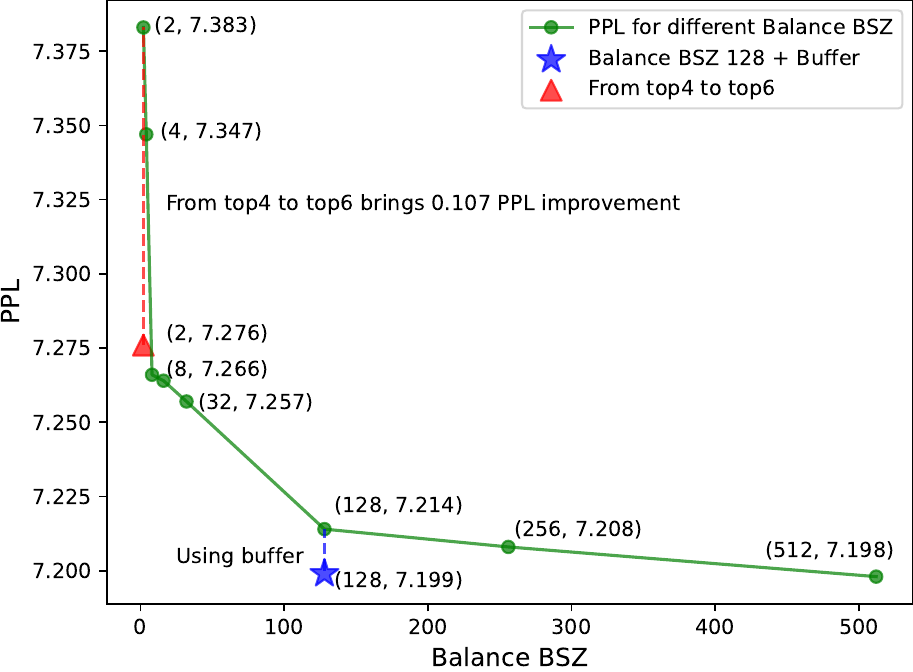}
\vskip  -0.1in
\caption{The performance of MoE-3.4A0.6B trained on 400B tokens with different Balance BSZ.}
\label{fig:scaling-batch}
\end{center}
\vskip  -0.15in
\end{wrapfigure}

\paragraph{Model performance increasing with Balance BSZ.} 
To further illustrate the impact of different Balance BSZes on the model performance, we controlled the micro-batch size, synchronization scope, and number of devices in training the 3.4A0.6B model on 400B tokens, and plotted the results from a Balance BSZ of 2 (micro-batch size 2, without any synchronization for expert selection frequency) to 512, as shown in Fig.~\ref{fig:scaling-batch}. 
As the Balance BSZ increases, the test PPL consistently decreases, \textbf{with an overall decrease of 0.185 from 2 to 512}. 
It is also noticeable that the improvement rate slows down after increasing to 128, and the result of adding the buffer is very close to that of 512. 
\textit{This indicates that synchronization and buffer mechanisms can bring significant improvements compared to micro-batch in MoE training across various computing node scales.}
Additionally, we supplemented experiments by increasing the activation from top4 experts to top6 experts under the micro-batch condition and found that the improvement brought by a 50\% increase in activated expert FLOPs is even less than the improvement from increasing the Balance BSZ from 2 to 8.
Since the additional overhead brought by synchronization and buffer is much less than that brought by increasing FLOPs, this further demonstrates the efficiency of expanding the Balance BSZ.

\section{Analysis}
\label{sec:analysis}

% \begin{table}[h]
\begin{wraptable}{r}{0.5\columnwidth}
\centering
\vskip  -0.15in
\caption{Ablation of the number of tokens and distributional bias for computing LBL on MoE-3.4A0.6B .}
\vskip  -0.1in
\label{tab:num_vs_dist}
\resizebox{0.5\columnwidth}{!}{
\begin{tabular}{c|ccc}
\toprule
\textbf{LBL type} & \textbf{Hellaswag} & \textbf{MMLU} & \textbf{Avg PPL} \\
\midrule
\multicolumn{4}{c}{120B Tokens, Global Batch Size 512, Micro Batch Size 4} \\
\midrule
Micro & 62.81 & 41.63 & 8.167 \\
Global & \textbf{63.75} & \textbf{43.48} & \textbf{8.038} \\
Shuffle & 63.57 & 43.37 & 8.041 \\
\midrule
\multicolumn{4}{c}{400B Tokens, Global Batch Size 1024, Micro Batch Size 2} \\
\midrule
Micro & 67.22 & 48.77 & 7.383 \\
Global & 68.32 & \textbf{49.84} & \textbf{7.198} \\
Shuffle & \textbf{68.43} & 49.68 & 7.214 \\
\bottomrule
\end{tabular}
}
\vskip  -0.1in
% \end{table}
\end{wraptable}

\paragraph{Ablation Study on Token Numbers and Token Distributional Bias}
As aforementioned, the crucial factor for global-batch LBL to outperform micro-batch LBL is that the latter pushes the router to achieve sequence-level balanced expert utilization, which may be overly stringent and hurt the model performance.
However, another naive assumption is that the $\text{LBL}_{\text{global}}$ involves more tokens to estimate the expert selection frequency, thus reducing the variance and ameliorating the MoE training.
To verify, we introduce another setting: \textit{Shuffle} $\text{LBL}_{\text{micro}}$. Specifically, when calculating the LBL, we first synchronize the token-expert score matrix $G$ (with a shape of number of tokens $\times$ number of experts) in all parallel groups, where $G_{ij}=1$ if the token $i$ selects the expert $j$, otherwise $G_{ij}=0$.
Then, we randomly select a batch of tokens (without replacing) to calculate the expert selection frequency, where the batch size is equal to the micro-batch size.
In this setting, the random batch has the same token numbers as the micro-batch and identical token distribution as the global-batch, enabling us to tell the difference between these two confounders.
The results are shown in the Tab.~\ref{tab:num_vs_dist}.
We experiment with two settings: train 120B tokens with a micro-batch size of 4 and global-batch size of 512, and train 400B tokens with a micro-batch size of 2 and a global-batch size of 1024.
We observe that the \textit{Shuffle} $\text{LBL}_{\text{micro}}$ achieves similar performance as $\text{LBL}_{\text{global}}$, and sill significantly outperforming the  $\text{LBL}_{\text{micro}}$, verifying the motivation of our paper and the assumption about the improvement.

\begin{figure}[h!]
% \vskip  -0.2in
\begin{center}
\includegraphics[width=0.37\columnwidth]{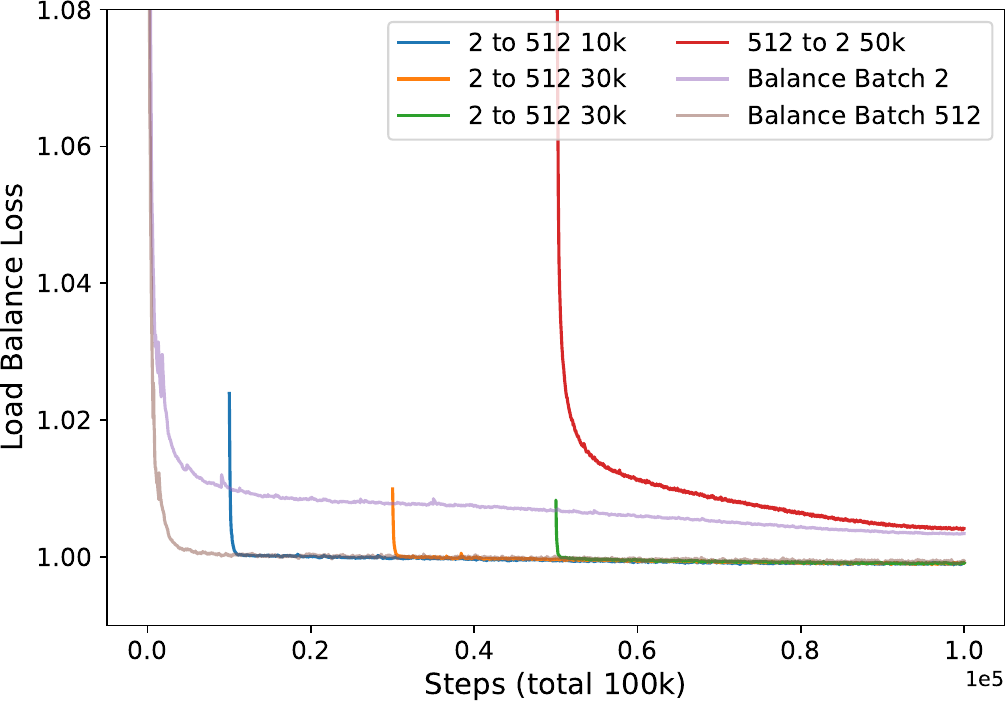}
\includegraphics[width=0.37\columnwidth]{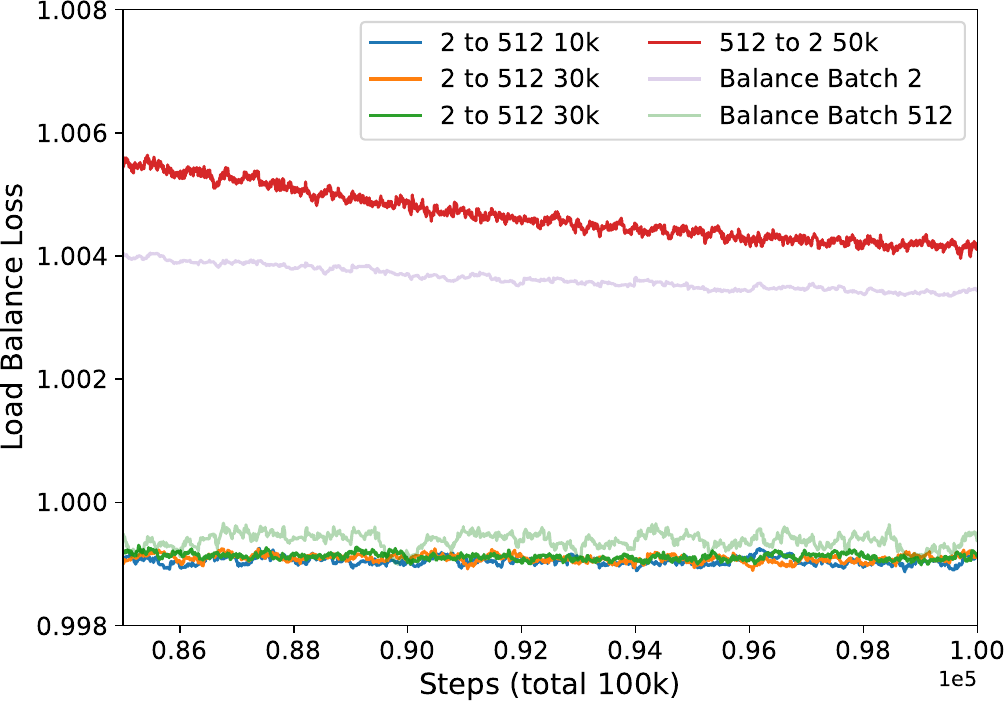}
\includegraphics[width=0.37\columnwidth]{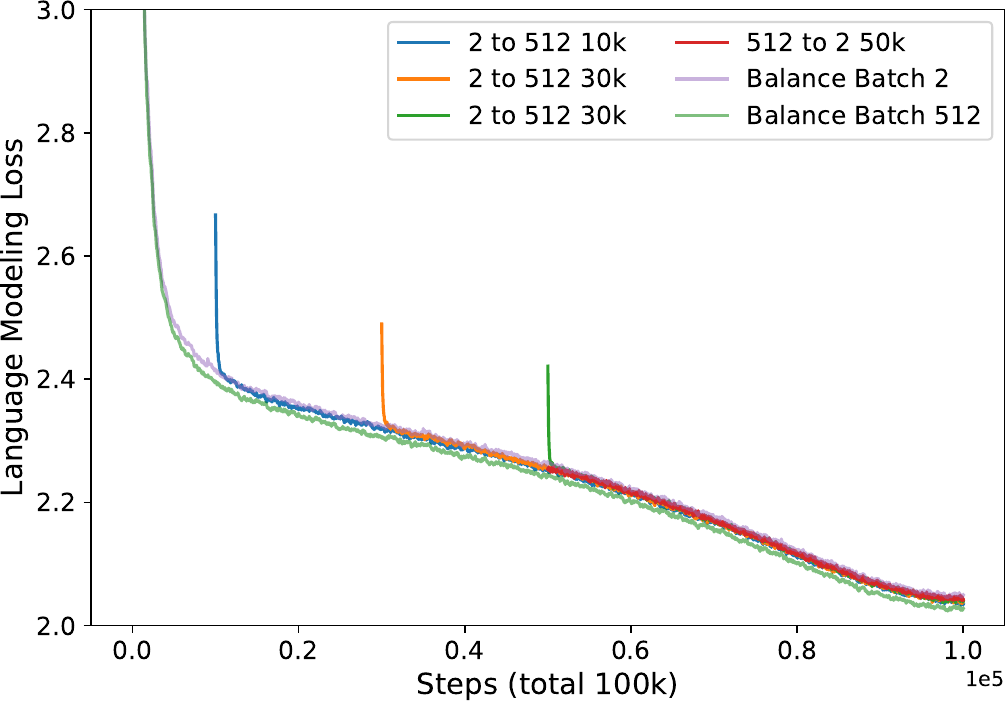}
\includegraphics[width=0.37\columnwidth]{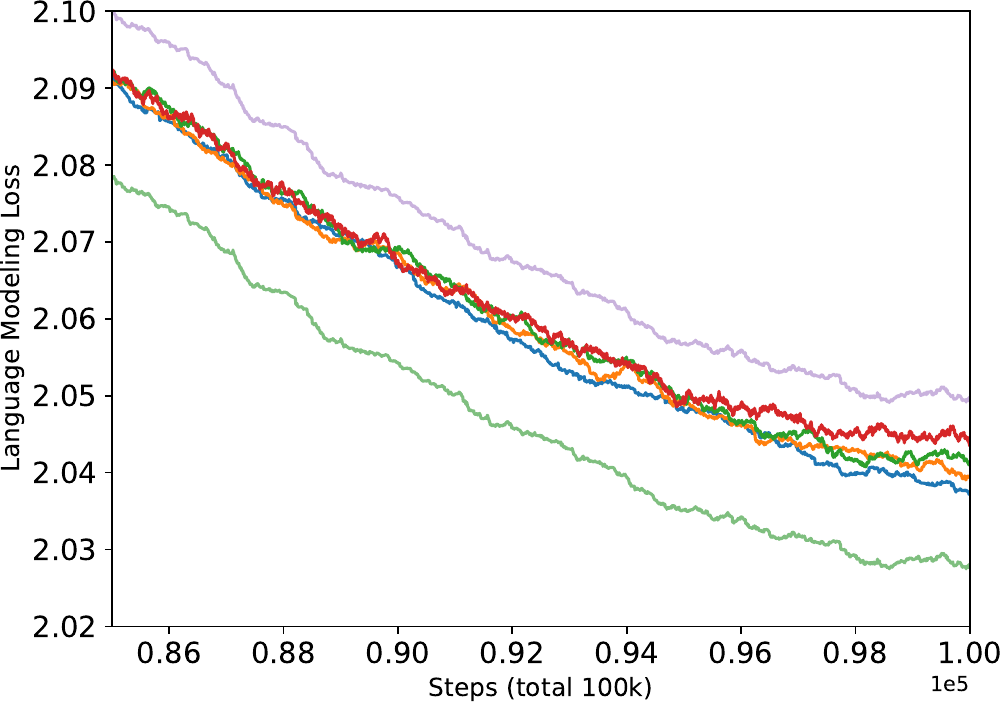}
\vskip  -0.1in
\caption{The LBL (Top) and language modeling loss (Bottom) curve for MoE-3.4A0.6B trained on 400B tokens under different Balance BSZ, with a zoom-in of the last 15k steps shown below.
In addition to the settings with Balance BSZ of 512 and 2, there are settings where the Balance BSZ changes from 2 to 512 at 10k, 30k, and 50k steps and a setting where it changes from 512 to 2 at the 50k step.}
\label{fig:lbl-curve-compare}
\end{center}
\vskip  -0.1in
\end{figure}

\paragraph{$\text{LBL}_{\text{global}}$ is a looser constraint than $\text{LBL}_{\text{micro}}$.}
Intuitively, global-batch balance is a looser constraint than micro-batch balance: the former requires only that tokens be evenly distributed across experts globally, while the latter demands uniform distribution within each micro-batch.
In Fig.~\ref{fig:lbl-curve-compare} (Top), we show the loss curves of the two methods using the same load balance weight for MoE-3.4A0.6B trained on 400B. 
Additionally, we add the results of switching from micro-batch balance to global-batch balance at 10k, 30k, and 50k training steps. 
It can be observed that (1) after switching to global-batch balance, the LBL rapidly decreases to a range close to that when the global-batch balance is used from scratch, and the final convergence trend is also similar. 
This is because transitioning from a tighter constraint (balance within a micro-batch) to a looser one (balance within a global-batch) is relatively easy.
(2)Moreover, if global batch balance is switched to micro-batch balance at the 50k step, the originally converged load balance first rises to a much higher range, then slowly decreases, and the final convergence result is still lower than that of micro-batch balance used from scratch. 
This indicates that transitioning from a looser constraint to a tighter one can significantly alter the convergence state.

\begin{wraptable}{r}{0.45\columnwidth}
\centering
\vskip  -0.15in
\caption{\footnotesize The impact of changing the Balance BSZ during training on the final results. 
Step indicates the step at which the Balance BSZ is switched.}
\vskip  -0.1in
\label{tab:change-balance}
\resizebox{0.38\textwidth}{!}{
\begin{tabular}{ccc}
\toprule
\textbf{Balance BSZ} & \textbf{Step (/100k)} & \textbf{PPL} \\
\midrule
2 & - & 7.383 \\
2$\rightarrow$512 & 50k & 7.322 \\
2$\rightarrow$512 & 30k & 7.297 \\
2$\rightarrow$512 & 10k & 7.283 \\
512 & - & \textbf{7.199} \\
512$\rightarrow$2 & 50k & 7.373 \\
\bottomrule
\end{tabular}
}
\vskip  -0.1in
\end{wraptable}

In Fig.~\ref{fig:lbl-curve-compare} (Bottom), we present the language modeling loss curves in the same setting, and the test PPL statistics for these settings are compiled in Tab.~\ref{tab:change-balance}.
It can be observed that (1) the loss of global-batch balance is over 0.02 lower than that of micro-batch balance, corresponding to the large performance gap between the two as shown in Tab.~\ref{tab:change-balance}. 
(2) Switching from micro-batch to global-batch balance results in performance improvements, with earlier switches yielding better outcomes. 
However, even the switch at the 10k step is notably inferior to training with global-batch balance from scratch. 
This aligns with existing findings that router choices tend to become fixed early in training~\citep{xue2024openmoe, muennighoff2024olmoe}: although increasing the Balance BSZ at any training stage can bring benefits, the router trained with micro-batch balance has already saturated very early, thus the gains from switching during training are limited.
(3) Switching from global-batch to micro-batch balance degrades performance, indicating that changes in expert selection during training can greatly affect model performance.

% \begin{table}[h]
\begin{wraptable}{r}{0.6\columnwidth}
\centering
\vskip  -0.1in
\caption{Results for different load balance weight.}
\vskip  -0.1in
\label{tab:compare-reduce-lbl}
\resizebox{0.6\textwidth}{!}{
\begin{tabular}{cc|ccc}
\toprule
    \textbf{Balance BSZ} & \textbf{LBL weight} & \textbf{Hellaswag} & \textbf{MMLU} & \textbf{Avg PPL} \\
\midrule
4 & 0.008 & 62.81 & 41.63 & 8.167 \\
4 & 0.004 & 62.95 & 42.13 & 8.154 \\
4 & 0.001 & 62.97 & 41.71 & 8.159 \\
512 & 0.008 & \textbf{63.75} & \textbf{43.48} & \textbf{8.038} \\
\bottomrule
\end{tabular}
}
\vskip  -0.15in
% \end{table}
\end{wraptable}

Since micro-batch balance is a tighter constraint than global-batch balance, we further test reducing the load balance weight of micro-batch balance in Tab~\ref{tab:compare-reduce-lbl}. 
It can be observed that appropriately reducing the LBL weight can slightly improve the model's performance within a certain range, but too small LBL weight leads to worse results. 
This may be due to the overly imbalanced distribution affecting the utilization of MoE model parameters. 
Moreover, the performance of micro-batch balance under various LBL weights is inferior to that of global-batch balance, further highlighting the differences between the two balancing methods.

\begin{wraptable}{r}{0.55\columnwidth}
\vskip  -0.15in
\centering
\caption{Performance and speed (seconds per iteration) in 43A6.6B setting. `128+buffer \& 8' means adding micro-batch balancing loss with Balance BSZ 8.}
\vskip  -0.1in
\label{tab:computation-time}
\resizebox{0.55\textwidth}{!}{
\begin{tabular}{c|ccc|c}
\toprule
\textbf{Balance BSZ} & \textbf{Hellaswag} & \textbf{MMLU} & \textbf{Avg PPL} & \textbf{Speed/s} \\
\midrule
8 & 75.20 & 54.98 & 5.862 & \textbf{1.55} \\
128+buffer & \textbf{75.94} & \textbf{57.30} & \textbf{5.779} & 1.64 \\
128+buffer \& 8 & 75.87 & 57.00 & 5.795 & 1.59 \\
\bottomrule
\end{tabular}
}
\vskip  -0.12in
\end{wraptable}

\paragraph{The computation cost and efficiency of global-batch balance.}
Because a dropless strategy is employed, the FLOPs calculation is identical across different methods. However, due to differences in local balance conditions, methods using global-batch balance may experience local computational imbalance. To address this, we recorded the speed and results of micro-batch balance and global-batch balance during the training of the 43A6.6B model in Tab.~\ref{tab:computation-time}. 
(1) It can be seen that the speed using global-batch balance (1.64 s/iteration) is 5.8\% slower than micro-batch balance (1.55 s/iteration). Further analysis revealed that about 1\% of this slowdown is due to communication overhead within all data parallel groups, the remainder mainly due to local expert load imbalance under the dropless strategy. 
Drawing inspiration from sequence-level LBL, we introduced a very low weight (1\% of the global-batch weight) micro-batch balancing loss into the global-batch balance at the 20k step and continued training the model. 
We found that (2) adding a small amount of micro-batch balancing loss increased the speed to 1.59 s/iteration (2.6\% slower than the baseline) with only a minimal decrease in performance. 
\textit{It should be noted that since the computation of LBL is independent from other parts of the network and takes very little time, it can be overlapped to further reduce the efficiency gap to within 2\%.}

\begin{figure*}[t!]
\begin{center}
\vskip  -0.2in
\centerline{\includegraphics[width=0.95\textwidth]{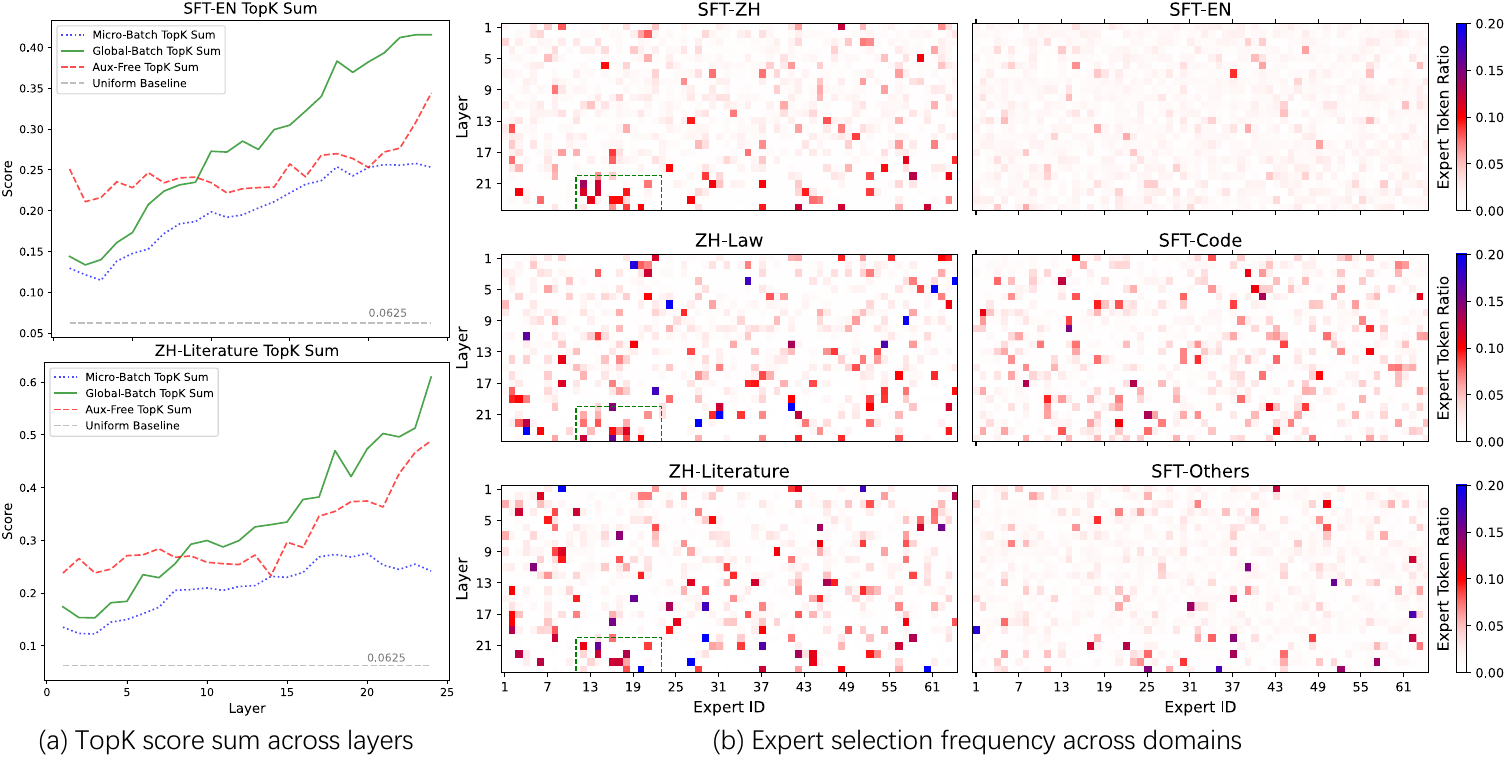}}
\vskip  -0.1in
\caption{The topK score sums across layers (a), and the distribution of high-frequency experts on different domains for models using global-batch balance (b). The topK sum of global-batch balance is higher than other methods and shows a similar distribution of high-frequency experts on closer domains.}
  \label{fig:topk_sum_select_frequency}
  \end{center}
  \vskip  -0.3in
\end{figure*}

\paragraph{Global batch balance brings interpretable specialization.}
In this section, we further analyze the specialization of models using global-batch balance. 
In Fig.~\ref{fig:topk_sum_select_frequency} (a), we record the scores assigned to each expert by tokens across different domains and calculate the average of the topK score sums. 
When all experts are assigned identity scores, the topK sum is indicated by the \textcolor{gray}{gray dashed line}.
We can observe: (1) Models using global-batch balance have a higher topK sum at each layer. 
Since the LBL and z-loss in MoE encourage routing scores to be uniform, while only the language modelling loss encourages an increase in routing scores, this suggests that \textit{under the global-batch balance, routing is more aligned with the language modelling task}. 
(2) Models using global-batch balance have a larger topK sum in domains where expert selection is more concentrated. 
For example, in Fig.~\ref{fig:topk_sum_select_frequency} (b), the high-frequency experts in ZH-Literature are more than those of SFT-EN, especially in layers 17 to 24. 
Correspondingly, in Figure 2(a), the topK sum of ZH-Literature in layers 17 to 24 is higher than that of SFT-EN. 
(3) Models using micro-batch balance have lower topK sums, with little difference across domains, which corresponds to the existing work that current MoE routing is uncertain~\citep{wu2024gw}. 
(4) Under global-batch balance, the topK sum of using aux loss free is smaller than that of LBL, but higher than micro-batch balance. \textit{This further illustrates that expert specialization promotes the concentration of expert scores.}

In Fig.~\ref{fig:topk_sum_select_frequency} (b), we compare the distribution of high-frequency experts across domains. 
We observed that several Chinese domains (SFT-ZH, ZH-Law, ZH-Literature) have many similar high-frequency experts (indicated by the dashed box). 
Moreover, although both Chinese-related domains and SFT-Code have high-frequency activated experts, these experts hardly overlap. 
For domains with more general content (such as SFT-EN), there are fewer instances of individual experts being highly activated.

\section{Related works}

\paragraph{Load Balancing}
\citet{DBLP:conf/iclr/ShazeerMMDLHD17} introduce a topK-controlled sparse activation mechanism in MoE~\citep{DBLP:conf/icnn/SzymanskiL93}, which tends to elect only a few experts for updates during training without constraints. 
Although LBL can alleviate this issue, overly strict constraints may also affect model performance. 
Expert Choice Routing~\citep{zhou2022mixture} achieves load balance naturally by allowing each expert to select tokens based on its load capacity. 
However, it uses the routing score information of the entire sequence when allocating tokens, making it non-causal for decoder-only models during inference.
Although subsequent work has found ways to add extra routers and training phases to address this, it has only been demonstrated when the number of experts equals 2~\citep{raposo2024mixture}.

\citet{wang2024auxiliary} argue that the load balance loss, which is not entirely consistent with the language modelling loss, can impact model performance. 
Therefore, they propose adding a bias term updated based on expert selection frequency to balance expert selection without changing routing scores. 
However, they don't emphasize whether the expert selection frequency is calculated based on micro-batch or global-batch. 
The subsequent work deepseek-v3~\citep{liu2024deepseek}, concurrent with ours, highlights that the expert selection frequency in Aux Loss Free is based on `the whole batch of each training step' and discusses the results of using batch-wise load balance loss and auxiliary free method, also finding that the two methods yield similar results. 
In this work, we propose synchronizing expert selection and buffering methods that can be easily integrated into existing MoE frameworks, leading to improvements under various computational configurations. 
Our work also provides a detailed analysis of Balance BSZ's impact on performance and demonstrates that global-batch significantly improves performance by incorporating more diverse domain information. 
Additionally, we show that adding a small amount of micro-batch load balance while using global-batch balance can maintain model performance while reducing latency from local imbalance.
Another concurrent work, Minimax-01~\citep{li2025minimax}, synchronizes expert select frequency within expert parallel groups, primarily aiming to reduce the drop rate of experts when using drop strategies~\citep{DBLP:journals/jmlr/FedusZS22}, without focusing on the impact of different Balance BSZ.

GRIN~\citep{liu2024gringradientinformedmoe} proposes Global Load Balance Loss Adaptations. 
However, the it mainly introduces this balance method as an advantage of the training framework without employing expert parallelism. 
GRIN does not present more motivation for using global load balance. Additionally, it does not show the effects of using global load balance independently and emphasizes the importance and properties of global load balance.

\paragraph{Expert Specialization} Initially, MoE is designed to \textit{devide and conquer}, allowing different experts to specialize strongly for efficient parameter utilization~\citep{DBLP:conf/icnn/SzymanskiL93,qiu2024unlockingemergentmodularitylarge}.
\citet{DBLP:conf/iclr/ShazeerMMDLHD17} introduce topK-based sparse activation mechanism to scale up model parameters. 
With the tight micro-batch balance, most MoE models~\citep{jiang2024mixtral}, including multimodal MoEs~\citep{lin2024moellavamixtureexpertslarge,chameleonteam2024chameleonmixedmodalearlyfusionfoundation}, have not exhibited domain-level specialization.
Lory~\citep{zhong2024loryfullydifferentiablemixtureofexperts} addresses this issue by calculating expert merge scores for each sequence and merging all experts into a single expert before computing the corresponding sequence.
This changes the sparse activation mechanism of MoE and avoids the imbalance issue.
% Additionally, Lory emphasizes packing similar sentences into a sequence to encourage expert specialization. 
Although Lory shows better improvements and specialization, its complex mechanism poses challenges for large-scale training.
OLMoE~\citep{muennighoff2024olmoeopenmixtureofexpertslanguage} analyzes the activation of experts across different domains and observes more pronounced specialization compared to Mixtral-8$\times$7B~\citep{jiang2024mixtral}. However, it does not provide a detailed discussion of the factors influencing specialization.

\section{Conclusion}
In this work, we identify that the LBL in mainstream MoE training frameworks has degraded into micro-batch balance in the era of LLMs, which imposes an overly tight constraint on routing decisions. 
This constraint not only limits expert specialization but also negatively impacts model performance. 
To address this issue, we propose methods based on synchronization and buffering to relax micro-batch balance to global-batch balance, a constraint more conducive to model specialization. 
We have validated the effectiveness of these methods across models of various sizes. 
Through a detailed analysis of expert selection under global-batch balance, we observe that it enables domain-level and interpretable characteristics. 
We hope that adopting the global-batch balance method will facilitate the development of more performant and interpretable MoE-based LLMs.

\section*{Limitations}

This paper primarily focuses on analyzing the impact of micro-batch LBL on LLMs during the pre-training stage. 
It does not further investigate its effects during fine-tuning or in the vision and multi-modality domains. 
Our analysis of specialization is mainly centred on the selection frequency across different domains without conducting more rigorous validation. 
Relaxing micro-batch LBL can introduce some latency. 
Future work could consider including more diverse sequences within each micro-batch to mitigate this local imbalance issue.

\bibliography{custom}
\bibliographystyle{colm2024_conference}

% \appendix

%\section{Appendix}
%\label{sec:appendix}

% \subsection{More Related Works}

% \paragraph{Expert Specialization}

\end{document}